\documentclass[runningheads]{llncs}
\usepackage{booktabs}
\usepackage[T1]{fontenc}
\usepackage{graphicx,verbatim}
\usepackage{amssymb}
\usepackage{pdfpages}
\usepackage{makecell}
\usepackage{caption}
\usepackage{amsmath}
\usepackage{adjustbox}
\usepackage{xcolor}
\usepackage{hyperref}

\begin{document}
%
\title{HASD: Hierarchical Adaption for Pathology Slide-Level Domain-Shift}

\author{
Jingsong Liu\textsuperscript{‡}\inst{1,3} \and 
Han Li\textsuperscript{‡}\inst{1,2} \and 
Chen Yang\inst{2,4} \and
Michael Deutges\inst{2,4} \and
Ario Sadafi\inst{2,4} \and
Xin You\inst{2} \and
Katharina Breininger\inst{5} \and
Nassir Navab\inst{2,3} \and
Peter J. Schüffler\inst{1,3} \thanks{Corresponding author: \email{peter.schueffler@tum.de}}
}

\institute{
Institute of Pathology, Technical University of Munich, TUM School of Medicine and Health, Munich, Germany
\and
Computer Aided Medical Procedures (CAMP), TU Munich, Munich, Germany 
\and
Munich Center for Machine Learning (MCML), Munich, Germany
\and
Institute of AI for Health, Computational Health Center, Helmholtz Munich, Munich, Germany 
\and
Center for AI and Data Science (CAIDAS), Julius-Maximilians-Universität Würzburg, Würzburg, Germany
}

\authorrunning{Liu et al.}
\maketitle
\footnotetext[1]{\textsuperscript{‡}These authors contributed equally to this work.}  

\begin{abstract}

Domain shift is a critical problem for artificial intelligence (AI) in pathology as it is heavily influenced by center-specific conditions.  Current pathology domain adaptation methods focus on image patches rather than whole-slide images (WSI), thus failing to capture global WSI features required in typical clinical scenarios.
In this work, we address the challenges of slide-level domain shift by proposing a Hierarchical Adaptation framework for Slide-level Domain-shift (HASD). HASD achieves multi-scale feature consistency and computationally efficient slide-level domain adaptation through two key components: (1) a hierarchical adaptation framework that integrates a Domain-level Alignment Solver for feature alignment, a Slide-level Geometric Invariance Regularization to preserve the morphological structure, and a Patch-level Attention Consistency Regularization to maintain local critical diagnostic cues; and (2) a prototype selection mechanism that reduces computational overhead. We validate our method on two slide-level tasks across five datasets, achieving a 4.1\% AUROC improvement in a Breast Cancer HER2 Grading cohort and a 3.9\% C-index gain in a UCEC survival prediction cohort. 
Our method provides a practical and reliable slide-level domain adaption solution for pathology institutions, minimizing both computational and annotation costs. Code is available at \href{https://github.com/TumVink/HASD}{https://github.com/TumVink/HASD}.

\end{abstract}

\section{Introduction}
\begin{figure}
\begin{center}
\includegraphics[width=\textwidth]{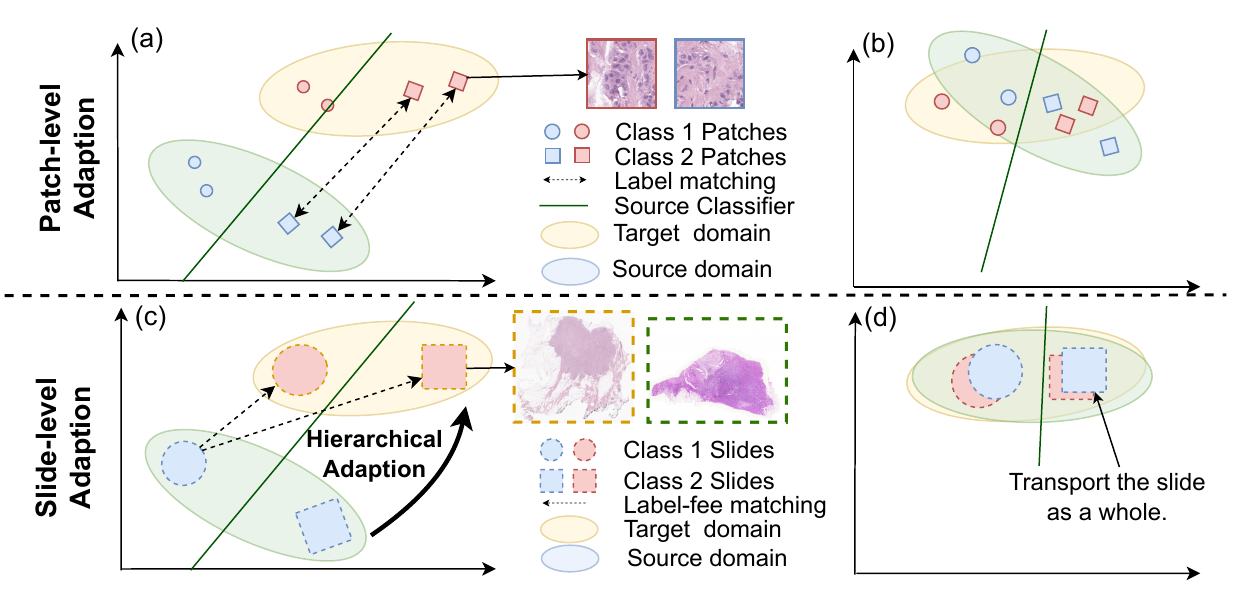}
\end{center}
\caption{(a)(b): Existing methods rely on patch-level labels from the source domain and pseudo labels from the target domain for patch-level domain adaptation. (c)(d): In slide-level tasks, patches lack individual labels. We propose a \textbf{hierarchical} adaption framework that leverages \textbf{domain-level} alignment solver with \textbf{slide-level} structure and \textbf{patch-level} diagnostic features preserved.}  \label{Figure 1}
\end{figure}

Digital pathology has experienced significant attention in recent years, driven by the widespread adoption of whole-slide imaging (WSI) scanners and rapid progress in artificial intelligence (AI) research~\cite{ffd}. This trend has facilitated notable advancements in many AI-based pathology tasks \cite{yin2024histosyn,fan2024pathmamba}. 
However, a major challenge in pathology is \textbf{domain shift}, arising from differences in pathology centers, including \textbf{imaging variances} (varying staining protocols, digitization procedures), \textbf{label discrepancies} (patient populations), and other factors. Domain shift can downgrade model performances when transferring a model to a different clinical site, potentially compromising patient safety \cite{unrobust,DAUD,xiong2023prior,xu2025nerf}.



Domain adaptation (DA) is a method to mitigate domain shift commonly used in histology image analysis, leveraging adversarial training or generative models \cite{DANN,scanner,DAUD,patho_gan,sharma2022mani,wang2021instance,falahkheirkhah2023domain}. By extracting domain-invariant features or synthesizing target-like samples, DA helps to align distributions between different datasets, improving model robustness and generalization. However, most methods focus on \textbf{patch-level adaption}, making them misaligned with real-world applications where clinical decisions require a global understanding of the slide (e.g., cancer grading or survival prediction) that patch-level adaptation methods fail to capture. An intuitive approach to extend patch-level DA methods to the slides-level is by decomposing slide-level tasks into many patch-level predictions~\cite{wang2022classification,yale_her2}, applying DA at the patch level, and then aggregating the slides-level results. However, this often requires extensive patch-level annotations from pathologists, which is very time-consuming. Another naive approach aggregates patch embeddings via simple integration techniques (e.g., mean pooling) \cite{enhancing1}. However, this treats all patches equally, failing to emphasize diagnostically relevant regions.  

Instead, we propose to realize\textbf{ slide-level} pathology DA by adopting the more adaptive attention-based aggregation (ABMIL).  
This presents two major challenges: \textbf{(1) The distortion of multi-scale feature distributions}: The success of ABMIL relies on the domain-level feature distributions, intra-slide patch relations, and patch-level attention patterns. Without a structured adaptation framework, DA can distort slide morphology and misalign critical diagnostic features. \textbf{(2) Computational and structural challenges:} The large number of patches per slide, varying  across different slides, introduces significant computational overhead, necessitating an efficient adaptation strategy. 

To solve these problems, 
we propose in this work an efficient Hierarchical Adaption framework for Slide-level Domain-shift (HASD).
Specifically, 
 HASD achieves multi-scale feature consistency and computationally efficient DA through two key components (Fig. \ref{Figure 1}): \textbf{(1) Hierarchical Adaptation Framework}: \underline{Domain-level} Alignment Solver, where an entropic Sinkhorn-Knopp solver \cite{sink} ensures effective alignment of feature distributions between domains; \underline{Slide-level} Geometric Invariance Regularization, ensuring that slides are adapted as a whole without structural distortions; \underline{Patch-level} Attention Consistency Regularization, which stabilizes domain adaptation by ensuring critical diagnostic features remain consistently focused across domains. \textbf{(2) Efficient Prototype Selection}: To mitigate the computational burden of the HASD, we select the $K$ most informative prototypes per slide, reducing redundancy while preserving essential slide-level information. Patch-level features are extracted using a pre-trained foundation model (UNI) \cite{uni}, ensuring robust representation learning. 

We demonstrate our method's DA capabilities in addressing imaging variances and label discrepancies with two slide-level tasks across five datasets. Compared to SOTA methods, our hierarchy adaption framework achieves an average \textbf{4.1\% } \textbf{AUROC} gain for Breast Cancer Grading and a \textbf{3.9\%} \textbf{C-index} gain in UCEC Survival Prediction task, \textbf{without} requiring additional pathologist annotations.
As such, our method provides a practical solution for pathology institutions seeking to transfer models from a source center to a target center while addressing domain shift, ensuring reliable adaptation with minimal computational overhead and annotation costs for slide-level tasks.

\section{Methods}

\begin{figure}[htbp]
\centering
\includegraphics[width=\textwidth]{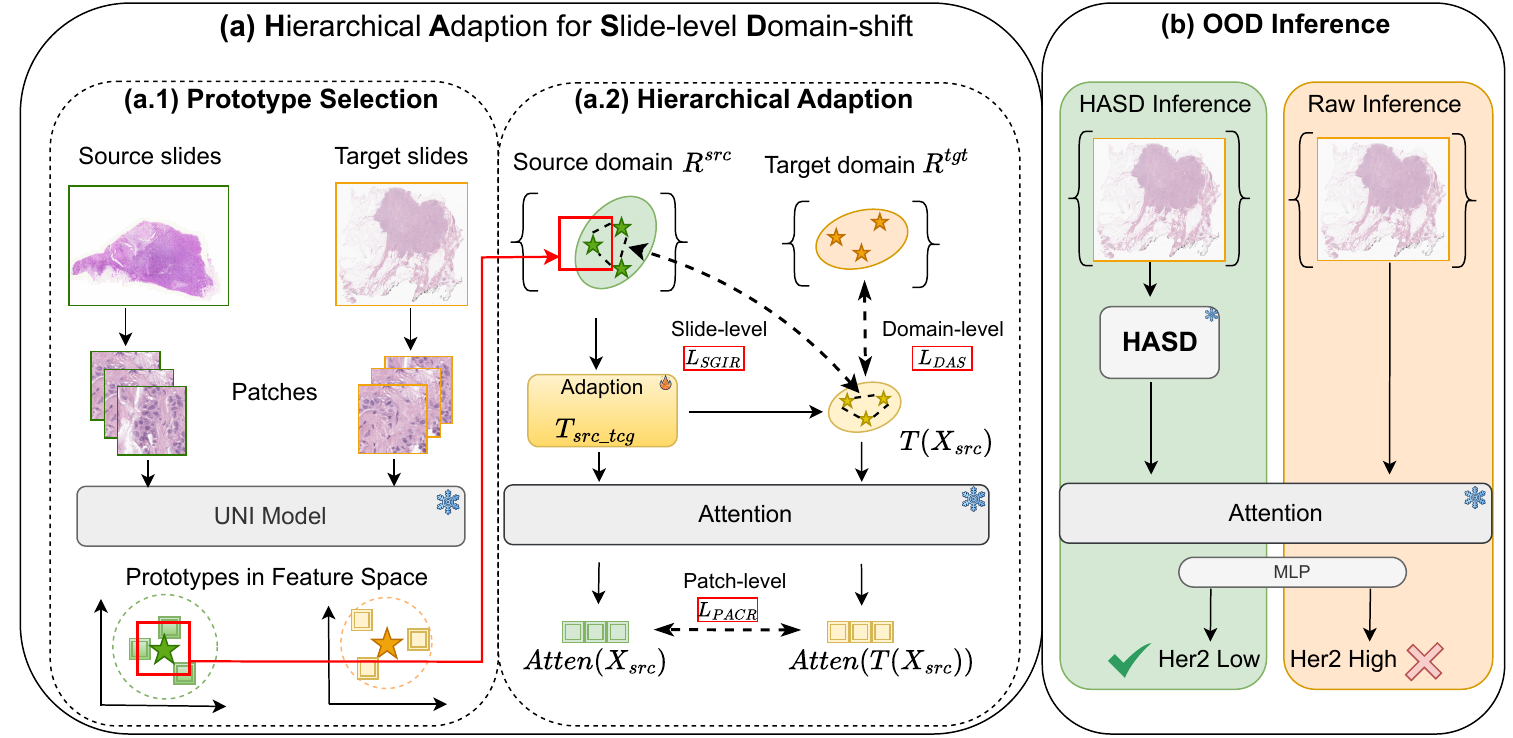} 
\caption{ HASD is designed for scenarios where UNI model extracts the patch features from both centers, and the attention aggregation with the MLP classifier are trained on the source domain. (a): The training process of HASD includes the selection of label-free prototypes in each slide and domain adaption in the hierarchical framework, including domain-, slide- and patch-level regularization. (b): Inference with or without HASD.} \label{Figure 2}
\end{figure}

\subsection{Problem Definition} \label{subsec:problem-definition}
\subsubsection{Multiple Instance Learning.}In a slide-level task with a cohort of $N$ slides, each slide is treated as a bag $B$ containing a different number of $P$ patches
extracted by a pretrained encoder into vectors $\{r_1,r_2,...,r_p\} \in \mathbb{R}^M$. Under the MIL assumption, each slide $B$ carries a single label $y$ (e.g., a cancer diagnosis), while individual patches lack direct labels. ABMIL \cite{abmil} learns an attention mechanism $Atten(\cdot)$ to aggregate the set of patch features into a slide-level prediction $\hat{y}$.
\subsubsection{Slide-Level Domain Shift.} We consider two domains: a source domain $\mathbf{R}^{src}$ where the model is trained, and a target domain $\mathbf{R}^{tgt}$ where the model is tested. Slide-level domain shift
cause the distribution of features in $\mathbf{R}^{src}$ to differ from that in $\mathbf{R}^{tgt}$. Formally, we represent the source and target domains as:
$$
\mathbf{R}^{src} = \{\mathbf{r}_{1,1}^{src},\, \mathbf{r}_{1,2}^{src},
\ldots,\, \mathbf{r}_{n,p}^{src}\},
\quad
\mathbf{R}^{tgt} = \{\mathbf{r}_{1,1}^{tgt},\, \mathbf{r}_{1,2}^{tgt},
\ldots,\, \mathbf{r}_{n,p}^{tgt}\},
$$
where $\mathbf{r}_{n,p}^{src}$ denotes the $p^{th}$ patch vector of the $n^{th}$ slide. 

\subsubsection{Optimal Transport} (OT) \cite{OT} provides a solution for learning a mapping between $\mathbf{R}^{src}$ and $\mathbf{R}^{tgt}$ while minimizing the overall cost:
$
\gamma\, c~\!(\mathbf{R}^{src},\, {\mathbf{R}^{tgt}}),
$
with $c(\cdot,\cdot)$ measures the feature distance (e.g., cosine similarity) and $\gamma$  matches samples from source to target. 

Traditional OT is not well-suited for pathology slide-level domain adaptation, as it incurs significant computational overhead \cite{yeaton2022hierarchical,xu2023multimodal} and treats each patch as an isolated point, thereby ignoring the structured slide morphology and failing to preserve multi-scale feature consistency \cite{OT_trend}. In the following section, we introduce our method to address these challenges.


\subsection{Hierarchical Alignment Framework}
\subsubsection{Domain-level Alignment Solver (DAS).}
Instead of directly using OT for domain adaptation, we propose optimizing a transformation function $T$ that maps $\mathbf{R}^{src}$ into a transition domain while minimizing the transport costs between $T(\mathbf{R}^{src})$ and $\mathbf{R}^{tgt}$. To efficiently compute the transport costs, we adopt a Sinkhorn–Knopp (SK) solver \cite{sink} by introducing an entropic regularization term $\mathcal{H}(\gamma)$.
Besides image variance, label discrepancy also poses a further challenge in slide-level domain adaptation.  To this end, we further introduce a partial mass relaxation $\mathcal{R}_{\text{partial}}(\gamma)$, allowing for samples to be excluded from matching by relaxing the marginal constraints via Kullback–Leibler (KL) divergence. Accordingly, we learn the transformation $T$ under the guidance of the total transport cost:
$$
\mathcal{L}_{\mathrm{DAS}}(T, \gamma)
\;=\;
\sum_{i,j}
\gamma_{ij}\, c ~\!\bigl(T\bigl(\mathbf{R}^{src}),\, \mathbf{R}^{tgt})
\;+\;
\varepsilon \mathcal{H}(\gamma)) + 
\tau \mathcal{R}_{\text{partial}}(\gamma),
$$
where $\gamma$ and $\tau$ control the strength of entropic regularization and partial relaxation, respectively. In our method, $\gamma$ is fixed as $0.001$, and we only activate the partial mass relaxation term (i.e. set $\tau \not = 0$) when label prevalence discrepancies occur across centers. In such cases (Fig. \ref{tsne} for an example), we refer to the resulting approach as \textbf{Partial DAS}. 

\subsubsection{Slide-level Geometric Invariance Regularization (SGIR).}
While DAS allows us to handle domain-level disparities, it operates on each patch independently, and we risk losing the local relationship between patches that belong to the same slide. 
Therefore, rather than treating patches as isolated points during alignment, we treat the slide as a whole and incorporate a slide-level constraint to enforce structural consistency in feature space. For this, we define our SGIR loss as:
$$
\mathcal{L}_{\mathrm{SGIR}}(T)
\;=\;
\sum_{n}^{N} \bigl\| G(B_n^{src}) - G(T(B_n^{src})) \|_F, 
\quad \text{where} \quad  B_n^{src} = \{r_{n,1}^{src},\ldots, r_{n,p}^{src}\},
$$
where $ \| \cdot \|_F $ denotes the Frobenius norm and the Gram matrix $G(X)=XX^T$. By preserving patch pairwise distance within the same slide, we ensure the slide-level structural consistency.

\subsubsection{Patch-level Attention Consistency Regularization (PACR).}
While domain-level and slide-level constraints preserve global structure, the adaptation can be over-warped, causing clinically essential patch-level local cues to be overlooked \cite{OT_trend}. To address this, we introduce PACR to preserve the aggregator’s patch-level attention distributions learned on the source domain. Intuitively, a patch with high attention in the source domain should maintain high attention after transformation by $T$. Formally, we define
$$
\mathcal{L}_{\mathrm{PACR}}(T)
\;=\;
\sum_{n}^N \sum_{p}^P \bigl\|Atten(T(r_{n,p}^{src})) \;-\; Atten(r_{n,p}^{src})\bigr\|^{2}.
$$
We thus form the total optimization objective as:
$$
\mathcal{L}_{total}(T,\gamma) = \mathcal{L}_{DAS}(T,\gamma) + \lambda_1 \mathcal{L}_{\mathrm{SGIR}}(T) + \lambda_2 \mathcal{L}_{\mathrm{PACR}}(T),
$$
with $\lambda_1$ and $\lambda_2$ balancing the domain alignment and regularization.

Overall, with our hierarchical adaption framework, we learn a transformation function $T: \mathbf{R}^{src} \rightarrow \mathbf{R}^{tgt},$ by minimizing the total objective function $L_{total}$, so that the model trained on $\mathbf{R}^{src}$ generalizes better to $\mathbf{R}^{tgt}$.


\subsection{Label-free Prototype Selection}
To be computationally efficient, we cluster each slide’s patches into $k$ groups and use the clusters' centroids as label-free prototypes (see ablation for $k$-selection) before applying HASD. This strategy significantly reduces complexity while preserving crucial slide-level information for alignment. An overview of our method is illustrated in Fig \ref{Figure 2}.

\section{Experiments and Results}
\newcommand{\greenup}{\textcolor{green}{$\uparrow$}}
\newcommand{\reddown}{\textcolor{red}{$\downarrow$}}
We validate our method with two slide-level tasks across five datasets:
\subsubsection{Breast Cancer HER2 Status.}
The prediction of the HER2 status in breast cancer from hematoxylin \& eosin (\textbf{H\&E}) stained tumor tissue is a challenging slide-level task \cite{bci}.
We use AUROC to evaluate our method on multi-institutional breast cancer slides from three centers: Yale Hospital (\textbf{Yale}, n=192) \cite{yale_her2}, The Cancer Genome Atlas (\textbf{TCGA}, n=182) \cite{tcga}, and Technical University Munich hospital (\textbf{TUM}, n=77).  

\subsubsection{UCEC Survival Prediction.} Uterine corpus endometrial carcinoma (UCEC) is a significant form of gynecological cancer \cite{ucec}. Reliable survival prediction for personalized treatment planning is limited by domain shift across different centers. To evaluate our method, we employ UCEC cases from two independent centers: \textbf{TCGA-UCEC} (n=504) and \textbf{CPTAC-UCEC} (n=205) \cite{cptac}.
We use disease-specific survival (\textbf{DSS}) and the Concordance Index (\textbf{C-Index}) to evaluate performances. In this cohort, domain shift arises not only from image variances (scanners, standings, etc.) but also from\textbf{ label prevalence variances} (Fig. \ref{tsne}a), posing an additional challenge for adaptation.

We conduct the domain adaption experiments in a \textbf{cross-validation} manner: for each center, 80\%  data is used to train a source-domain  model, which is then evaluated on (1) the rest 20\% in-domain test splits (\textbf{ID}) or (2) the entirety (100\%) of each out-of-domain target datasets in the same cohort, completely  held-out from the source training split (\textbf{OOD}, denoted as $source \rightarrow target$).

\subsection{Quantitative Evaluation}
We exhaustively compare HASD in eight OOD and five ID setups with the following four existing methods:
(1) Attention Based Multi Instance Learning \textbf{(ABMIL)}. We include a native ABMIL approach trained on the source data and directly applied  to the target data. 
(2)\textbf{ Patch-level Stain Normalization.} Widely used to mitigate domain shift on patch-level. We apply the Reinhard stain normalization \cite{reinhard}, instead of Macenko due to its much higher runtime, to transform all target patches into the source domain’s color space.
(3) Multi Instance SimpleShot\textbf{ (MI-SimpleShot).} Handles domain shifts by constructing class prototypes from slide features obtained by mean pooling of patch-level features \cite{uni}.
(4)\textbf{ Supervised Contrastive Domain Adaptation (SCDA).} Builds upon MI-SimpleShot with supervised contrastive learning on the source domain to refine class prototypes \cite{enhancing1}.








\begin{figure}[htbp]
  \centering
  \begin{minipage}[t]{0.35\textwidth}
      \centering
        \captionof{table}{ID vs. OOD performance gap.}
    \begin{adjustbox}{width=1\columnwidth}
    \begin{tabular}[\linewidth]{c  c c }
         \multicolumn{3}{c}{Her2 Grading Task}\\
     \hline
        {Methods}&\makecell[c]{ID\\Average} & \makecell[c]{OOD\\Average}  \\
    \hline

     {MI-SimpleShot}&81.6&70.8(10.8\reddown)\\
    
    {SCDA}&{82.5}&72.0(10.5\reddown)\\
    {ABMIL}&85.8&75.8(10.0\reddown)\\
    
     {ReinhardNorm}&{85.3}&75.3(10.0\reddown)\\
    \hline
    {Ours}&{\textbf{86.1}}&\textbf{79.9}(\textbf{6.2}\reddown)\\
    \bottomrule
    \\
    \\

    \multicolumn{3}{c}{Survival Prediction Task}\\
     \hline
        {Methods}&\makecell[c]{ID\\Average} & \makecell[c]{OOD\\Average} \\
    \hline
     {MI-SimpleShot}&64.9&52.9(12.0\reddown)\\
    
    {SCDA}&{65.3}&54.7(10.6\reddown)\\
    {ABMIL}&68.1&57.5(10.6\reddown)\\
    
     {ReinhardNorm}&{\textbf{68.5}}&56.3(12.2\reddown)\\
    \hline
    {Ours}&{68.1}&{\textbf{61.4}(\textbf{6.7}\reddown)}\\
    \bottomrule

    \end{tabular}
    \end{adjustbox}
    \label{tab:results}
  \end{minipage}\hfill
  \begin{minipage}[t]{0.60\textwidth}
    \centering
        \captionof{figure}{OOD Performances on HER2 Grading (\textcolor{red}{red}) and Survival Prediction (\textcolor{cyan}{cyan}).}
    \vspace{0pt}
    \includegraphics[width=1.0\textwidth]{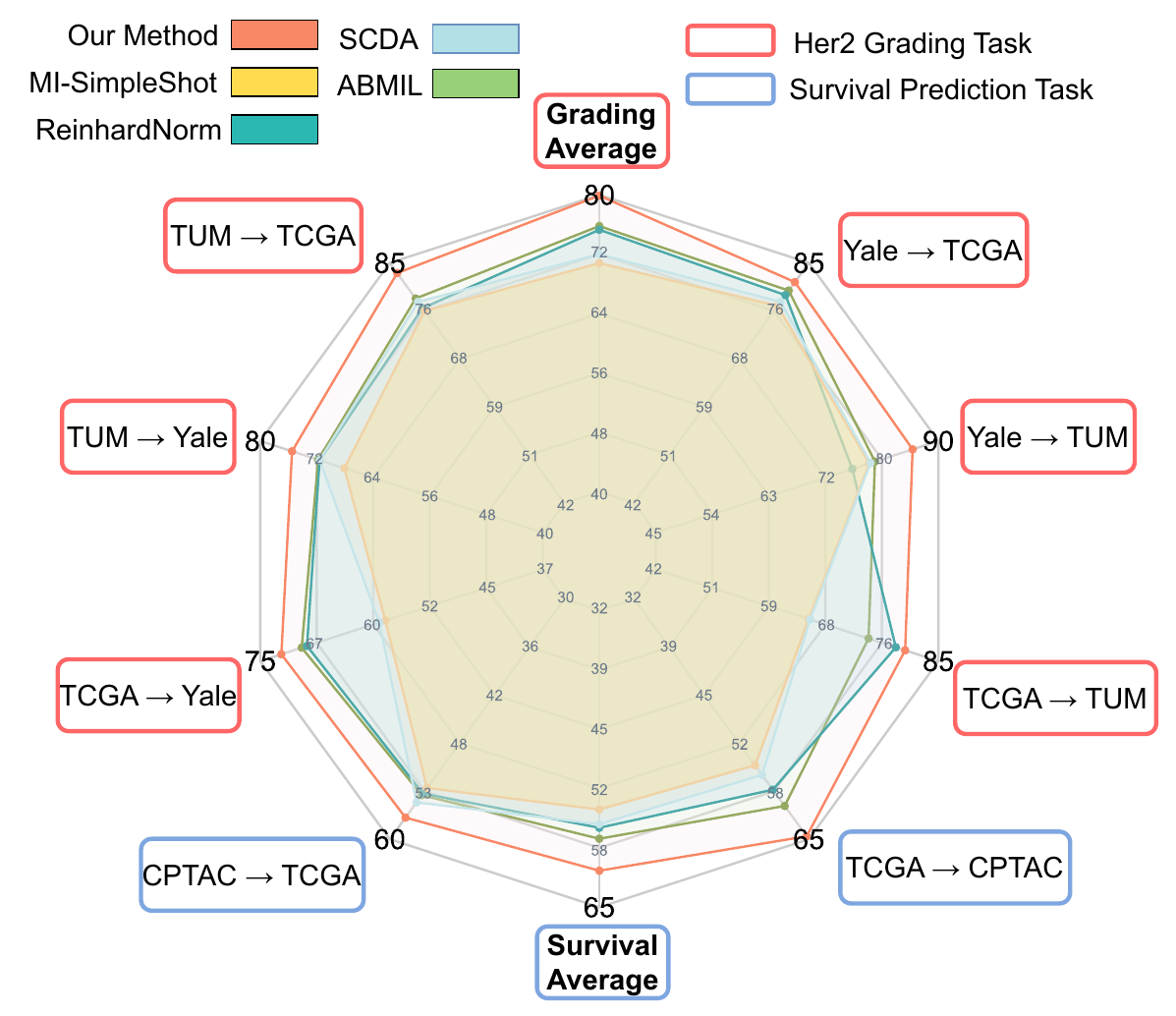}

    \label{fig:radar}
  \end{minipage}
\end{figure}

\subsubsection{Measuring Slide-level Domain Adaption.}
We report the performance of all evaluated methods under various OOD setups (see Fig. \ref{fig:radar}). In the HER2 Grading task, our method consistently outperforms others except for setup $TGCA \rightarrow TUM$, where stain normalization achieves the best result. Our model surpasses the next-best method by $\mathbf{6.0\%}$ \textbf{AUC in the most favorable case}, with an average improvement of $\mathbf{4.1\%}$ \textbf{across all setups}. For the Survival Prediction task, our method, enhanced by partial DAS to address label prevalence differences, consistently outperforms all baselines, achieving an average \textbf{CI improvement}  $\mathbf{3.9\%}$ over the next-best method.


\subsubsection{ID vs. OOD Gap Robustness Analysis.} 
Evaluating the ID vs. OOD gap is essential for assessing \textbf{domain shift robustness}. A smaller gap indicates better generalization. Table \ref{tab:results} summarizes the differences.
For HER2 Grading, MI-SimpleShot suffers the largest drop ($10.8\%$ AUC), while our method maintains the smallest gap ($6.2\%$ AUC) with the highest OOD score, demonstrating strong generalization.
In Survival Prediction, domain shift is further influenced by \textbf{label prevalence differences}. All other methods drops $> 10\%$ CI, while our method, aided by \textbf{partial DAS}, achieves the best OOD performance with a smaller gap ($6.7\%$ CI), proving its effectiveness.



\subsection{Ablation study} 
We summarize the impact of hierarchical components, including DAS, PACR, and SGIR, and the number of prototypes on both slide-level tasks in Table \ref{ablation_study}. Due to  GPU memory constraints, the prototype number is limited to $10$.

\newsavebox{\tablebox}
\begin{table}[t]
\centering
\caption{Ablation Study for Parameter Selection. } 


\label{ablation_study}
\begin{lrbox}{\tablebox}
\begin{tabular}{c c c c c c c c c } 
\toprule
\fontsize{8}{10}\selectfont
DAS&\makecell[c]{\# Prototype\\$k$}& \makecell[c]{PACR}&\makecell[c]{SGIR}  &\makecell[c]{Memory \\Usage (GB)}& \makecell[c]{ Grading\\Average} & \makecell[c]{Survial Prediction\\Average}\\ 
\hline
$\checkmark$&1&&&0.6&75.4&58.2\\ \hline
$\checkmark$&5&&&13.2&76.2(0.8\greenup)&58.6(0.4\greenup)\\ \hline
$\checkmark$&10&&&52.9&76.1(0.1\reddown)&59.1(0.5\greenup)\\ \hline
$\checkmark$&10&$\checkmark$&&52.9&79.6(3.5\greenup)&58.6(0.5\reddown)\\ \hline
$\checkmark$&10&$\checkmark$&$\checkmark$&53.0&79.9(0.3\greenup)&60.1(1.5\greenup)\\ \hline
Partial DAS&10&$\checkmark$&$\checkmark$&53.0&-&61.4(1.3\greenup)\\
\bottomrule
\end{tabular}

\end{lrbox}
\scalebox{0.8}{\usebox{\tablebox}}
\end{table}

\subsection{Quantitative Evaluation of Image \& Label Variance Alignment} \label{label_var}
\begin{figure}
\begin{center}
\includegraphics[width=1\textwidth]{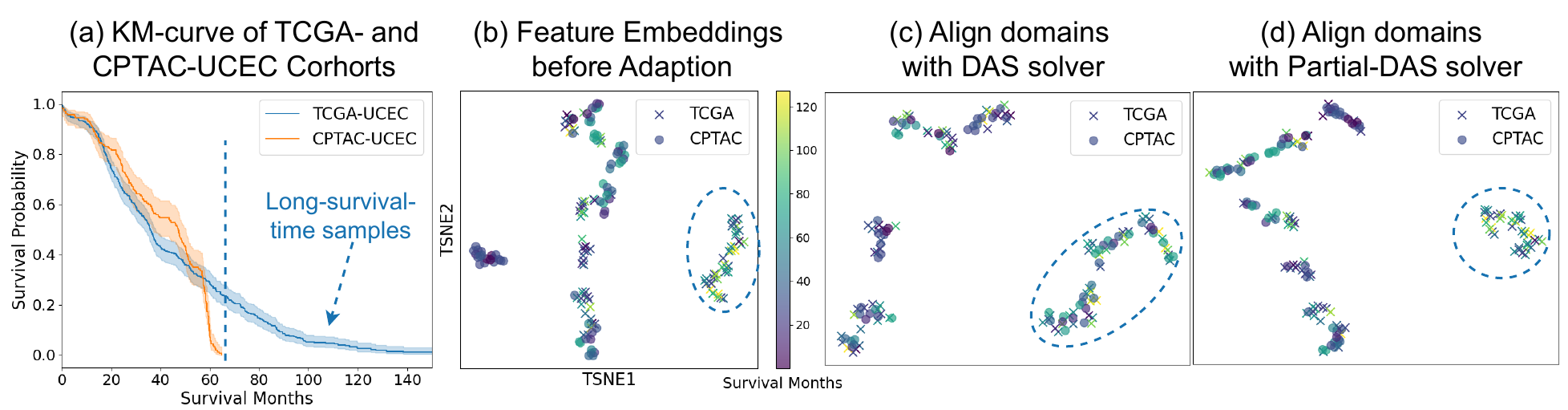}
\end{center}
\caption{T-SNE for the Survival task. (a) The TCGA data contain more long-surviving patients than CPTAC. (b) A clear domain shift exists between centers. (c) Balanced SB solver misaligns long-surviving TCGA samples with CPTAC, leading to poor adaptation. (d) Partial SB solver allows unmatched long-survival samples to remain unaligned, preserving clinical label relevance.}  \label{tsne}
\end{figure}

We illustrate how our method addresses \textbf{image variances} and alleviates \textbf{label prevalence variance} in t-SNE plots (Fig. \ref{tsne}) visualizing the aggregated slide-level features in the Survival Prediction task before and after adaptation. 


\section{Conclusion}
Domain shift is a critical problem for pathology AI as pathology data is heavily influenced by center-specific conditions. Current pathology domain adaptation methods focus on image patches rather than WSI, thus failing to capture global WSI features required in typical clinical scenarios. We, for the first time, introduced a slide-level domain adaption method HASD, which incorporates a hierarchical adaptation design to maintain multi-scale feature consistency and a prototype selection mechanism to achieve computationally efficiency. Our method outperformed conventional DA methods in two tasks (Her2 grading and Survival) by 4.1\% and 3.9\%, when models were trained and tested in different centers, while preserving in-domain performance. We believe that our research on slide-level domain adaptation will improve the generalizability of future clinical AI models. 


\bibliographystyle{splncs04}
\bibliography{ref.bib}

\begin{thebibliography}{10}
\providecommand{\url}[1]{\texttt{#1}}
\providecommand{\urlprefix}{URL }
\providecommand{\doi}[1]{https://doi.org/#1}

\bibitem{enhancing1}
Carretero, I., Meseguer, P., del Amor, R., Naranjo, V.: Enhancing whole slide image classification through supervised contrastive domain adaptation. arXiv preprint arXiv:2412.04260  (2024)

\bibitem{uni}
Chen, R.J., Ding, T., Lu, M.Y., Williamson, D.F., Jaume, G., Song, A.H., Chen, B., Zhang, A., Shao, D., Shaban, M., et~al.: Towards a general-purpose foundation model for computational pathology. Nature Medicine  \textbf{30}(3),  850--862 (2024)

\bibitem{sink}
Cuturi, M.: Sinkhorn distances: Lightspeed computation of optimal transport. Advances in neural information processing systems  \textbf{26} (2013)

\bibitem{cptac}
Edwards, N.J., Oberti, M., Thangudu, R.R., Cai, S., McGarvey, P.B., Jacob, S., Madhavan, S., Ketchum, K.A.: The cptac data portal: a resource for cancer proteomics research. Journal of proteome research  \textbf{14}(6),  2707--2713 (2015)

\bibitem{falahkheirkhah2023domain}
Falahkheirkhah, K., Lu, A., Alvarez-Melis, D., Huynh, G.: Domain adaptation using optimal transport for invariant learning using histopathology datasets. arXiv preprint arXiv:2303.02241  (2023)

\bibitem{fan2024pathmamba}
Fan, J., Lv, T., Di, Y., Li, L., Pan, X.: Pathmamba: Weakly supervised state space model for multi-class segmentation of pathology images. In: International Conference on Medical Image Computing and Computer-Assisted Intervention. pp. 500--509. Springer (2024)

\bibitem{yale_her2}
Farahmand, S., Fernandez, A.I., Ahmed, F.S., Rimm, D.L., Chuang, J.H., Reisenbichler, E., Zarringhalam, K.: Deep learning trained on hematoxylin and eosin tumor region of interest predicts her2 status and trastuzumab treatment response in her2+ breast cancer. Modern Pathology  \textbf{35}(1),  44--51 (2022)

\bibitem{DANN}
Ganin, Y., Ustinova, E., Ajakan, H., Germain, P., Larochelle, H., Laviolette, F., March, M., Lempitsky, V.: Domain-adversarial training of neural networks. Journal of machine learning research  \textbf{17}(59),  1--35 (2016)

\bibitem{scanner}
Ganz, J., Puget, C., Ammeling, J., Parlak, E., Kiupel, M., Bertram, C.A., Breininger, K., Klopfleisch, R., Aubreville, M.: Assessment of scanner domain shifts in deep multiple instance learning. In: BVM Workshop. pp. 137--142. Springer (2024)

\bibitem{abmil}
Ilse, M., Tomczak, J., Welling, M.: Attention-based deep multiple instance learning. In: International conference on machine learning. pp. 2127--2136. PMLR (2018)

\bibitem{unrobust}
de~Jong, E.D., Marcus, E., Teuwen, J.: Current pathology foundation models are unrobust to medical center differences. arXiv preprint arXiv:2501.18055  (2025)

\bibitem{bci}
Liu, S., Zhu, C., Xu, F., Jia, X., Shi, Z., Jin, M.: Bci: Breast cancer immunohistochemical image generation through pyramid pix2pix. In: Proceedings of the IEEE/CVF Conference on Computer Vision and Pattern Recognition (CVPR) Workshops. pp. 1815--1824 (June 2022)

\bibitem{DAUD}
Naranjo, V.: Domain adaptation for unsupervised cancer detection: An application for skin whole slides images from an interhospital dataset. In: International Conference on Medical Image Computing and Computer-Assisted Intervention. Springer (2024)

\bibitem{OT_trend}
Peyr{\'e}, G., Cuturi, M., et~al.: Computational optimal transport: With applications to data science. Foundations and Trends{\textregistered} in Machine Learning  \textbf{11}(5-6),  355--607 (2019)

\bibitem{ffd}
Pocevi{\v{c}}i{\=u}t{\.e}, M., Eilertsen, G., Garvin, S., Lundstr{\"o}m, C.: Detecting domain shift in multiple instance learning for digital pathology using fr{\'e}chet domain distance. In: International Conference on Medical Image Computing and Computer-Assisted Intervention. pp. 157--167. Springer (2023)

\bibitem{reinhard}
Reinhard, E., Adhikhmin, M., Gooch, B., Shirley, P.: Color transfer between images. IEEE Computer graphics and applications  \textbf{21}(5),  34--41 (2001)

\bibitem{patho_gan}
Reisenb{\"u}chler, D., Luttner, L., Schaadt, N.S., Feuerhake, F., Merhof, D.: Unsupervised latent stain adaptation for computational pathology. In: International Conference on Medical Image Computing and Computer-Assisted Intervention. pp. 755--765. Springer (2024)

\bibitem{sharma2022mani}
Sharma, Y., Syed, S., Brown, D.E.: Mani: Maximizing mutual information for nuclei cross-domain unsupervised segmentation. In: International Conference on Medical Image Computing and Computer-Assisted Intervention. pp. 345--355. Springer (2022)

\bibitem{ucec}
Somasegar, S., Bashi, A., Lang, S.M., Liao, C.I., Johnson, C., Darcy, K.M., Tian, C., Kapp, D.S., Chan, J.K.: Trends in uterine cancer mortality in the united states: a 50-year population-based analysis. Obstetrics \& Gynecology  \textbf{142}(4),  978--986 (2023)

\bibitem{OT}
Thorpe, M.: Introduction to optimal transport. Notes of Course at University of Cambridge  (2018)

\bibitem{wang2022classification}
Wang, P., Li, P., Li, Y., Xu, J., Jiang, M.: Classification of histopathological whole slide images based on multiple weighted semi-supervised domain adaptation. Biomedical Signal Processing and Control  \textbf{73},  103400 (2022)

\bibitem{wang2021instance}
Wang, Z., Zhu, X., Su, L., Meng, G., Zhang, J., Li, A., Wang, M.: Instance-aware feature alignment for cross-domain cell nuclei detection in histopathology images. In: Medical Image Computing and Computer Assisted Intervention--MICCAI 2021: 24th International Conference, Strasbourg, France, September 27--October 1, 2021, Proceedings, Part VIII 24. pp. 499--508. Springer (2021)

\bibitem{tcga}
Weinstein, J.N., Collisson, E.A., Mills, G.B., Shaw, K.R., Ozenberger, B.A., Ellrott, K., Shmulevich, I., Sander, C., Stuart, J.M.: The cancer genome atlas pan-cancer analysis project. Nature genetics  \textbf{45}(10),  1113--1120 (2013)

\bibitem{xiong2023prior}
Xiong, Y., Liu, J., Zaripova, K., Sharifzadeh, S., Keicher, M., Navab, N.: Prior-radgraphformer: A prior-knowledge-enhanced transformer for generating radiology graphs from x-rays. In: International Conference on Medical Image Computing and Computer-Assisted Intervention. pp. 54--63. Springer (2023)

\bibitem{xu2023multimodal}
Xu, Y., Chen, H.: Multimodal optimal transport-based co-attention transformer with global structure consistency for survival prediction. In: Proceedings of the IEEE/CVF international conference on computer vision. pp. 21241--21251 (2023)

\bibitem{xu2025nerf}
Xu, Z., Li, H., Sun, D., Li, Z., Li, Y., Kong, Q., Cheng, Z., Navab, N., Zhou, S.K.: Nerf-based cbct reconstruction needs normalization and initialization. arXiv preprint arXiv:2506.19742  (2025)

\bibitem{yeaton2022hierarchical}
Yeaton, A., Krishnan, R.G., Mieloszyk, R., Alvarez-Melis, D., Huynh, G.: Hierarchical optimal transport for comparing histopathology datasets. arXiv preprint arXiv:2204.08324  (2022)

\bibitem{yin2024histosyn}
Yin, C., Liu, S., Wong, V.W.S., Yuen, P.C.: Histosyn: Histomorphology-focused pathology image synthesis. In: International Conference on Medical Image Computing and Computer-Assisted Intervention. pp. 200--210. Springer (2024)

\end{thebibliography}
\end{document}